\documentclass{article}

\usepackage{arxiv}

\usepackage[utf8]{inputenc} % allow utf-8 input
\usepackage[T1]{fontenc}    % use 8-bit T1 fonts
\usepackage{hyperref}       % hyperlinks
\usepackage{url}            % simple URL typesetting
\usepackage{booktabs}       % professional-quality tables
\usepackage{amsfonts}       % blackboard math symbols
\usepackage{nicefrac}       % compact symbols for 1/2, etc.
\usepackage{microtype}      % microtypography
\usepackage{lipsum}		% Can be removed after putting your text content
\usepackage{graphicx}
\usepackage{natbib}
\usepackage{tabularx}
\usepackage{doi}

\usepackage{amsmath}
\usepackage{dsfont}
\usepackage{subfig}
\usepackage{xcolor}
\usepackage{appendix}
\usepackage{algorithm}
\usepackage{algpseudocode}

\usepackage{circledsteps}
\newcommand{\CCircled}[2][]{\begingroup
\pgfkeys{/csteps/.cd,inner color=black}%
\ifmmode
\Circled{$#2$}%
\else
\Circled{#2}%
\fi
\endgroup}

\title{TractOracle: towards an anatomically-informed reward function for RL-based tractography}

\author{ Antoine Théberge \\
	Department of Computer Science\\
	Faculty of Science \\
	  University of Sherbrooke \\
	\texttt{antoine.theberge@usherbrooke.ca} \\
        \And
	\And
	  Maxime Descoteaux \\
	Department of Computer Science\\
	Faculty of Science \\
	  University of Sherbrooke \\
	\texttt{maxime.descoteaux@usherbrooke.ca} \\
 	\And
	Pierre-Marc Jodoin \\
	Department of Computer Science\\
	Faculty of Science \\
	  University of Sherbrooke \\
	\texttt{pierre-marc.jodoin@usherbrooke.ca} \\
}

\renewcommand{\shorttitle}{TractOracle}

\hypersetup{
pdftitle={TractOracle},
pdfsubject={cs.LG},
pdfauthor={Antoine Théberge},
pdfkeywords={Tractography, Filtering, Difusion MRI},
}

\begin{document}
\maketitle

\begin{abstract}
    Reinforcement learning (RL)-based tractography is a competitive alternative to machine learning and classical tractography algorithms due to its high anatomical accuracy obtained without the need for any annotated data.  However, the reward functions so far used to train RL agents do not encapsulate anatomical knowledge which causes agents to generate spurious false positives tracts.  In this paper, we propose a new RL tractography system, \emph{TractOracle}, which relies on a reward network trained for streamline classification.  This network is used both as a reward function during training as well as a mean for stopping the tracking process early and thus reduce the number of false positive streamlines.  This makes our system a unique method that evaluates and reconstructs WM streamlines at the same time.  We report an improvement of true positive ratios by almost 20\%  and a reduction of 3x of false positive ratios on one dataset and an increase between 2x and 7x in the number true positive streamlines on another dataset.
\end{abstract}

% keywords can be removed

\section{Introduction}
\begin{figure}%
    \centering
    \subfloat[\centering]{{\includegraphics[height=3.0cm]{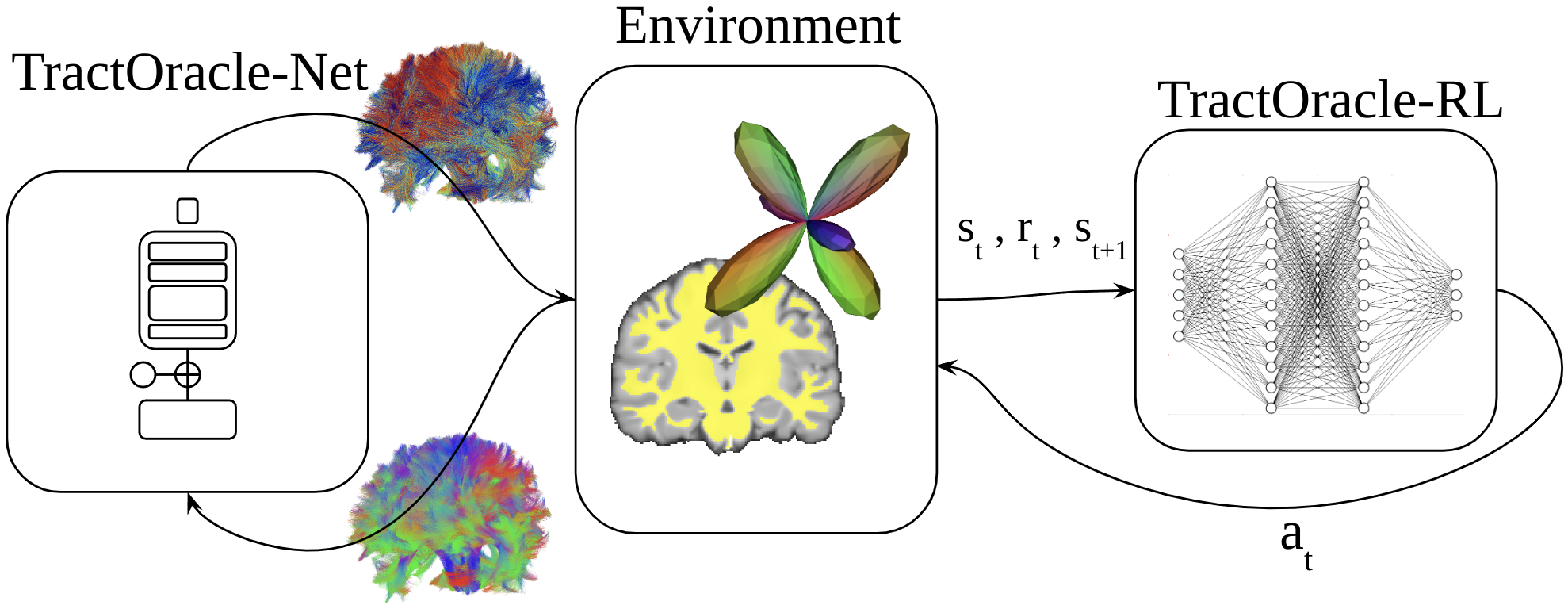} }}%
    % \qquad
    \subfloat[\centering]{{\includegraphics[width=4.0cm]{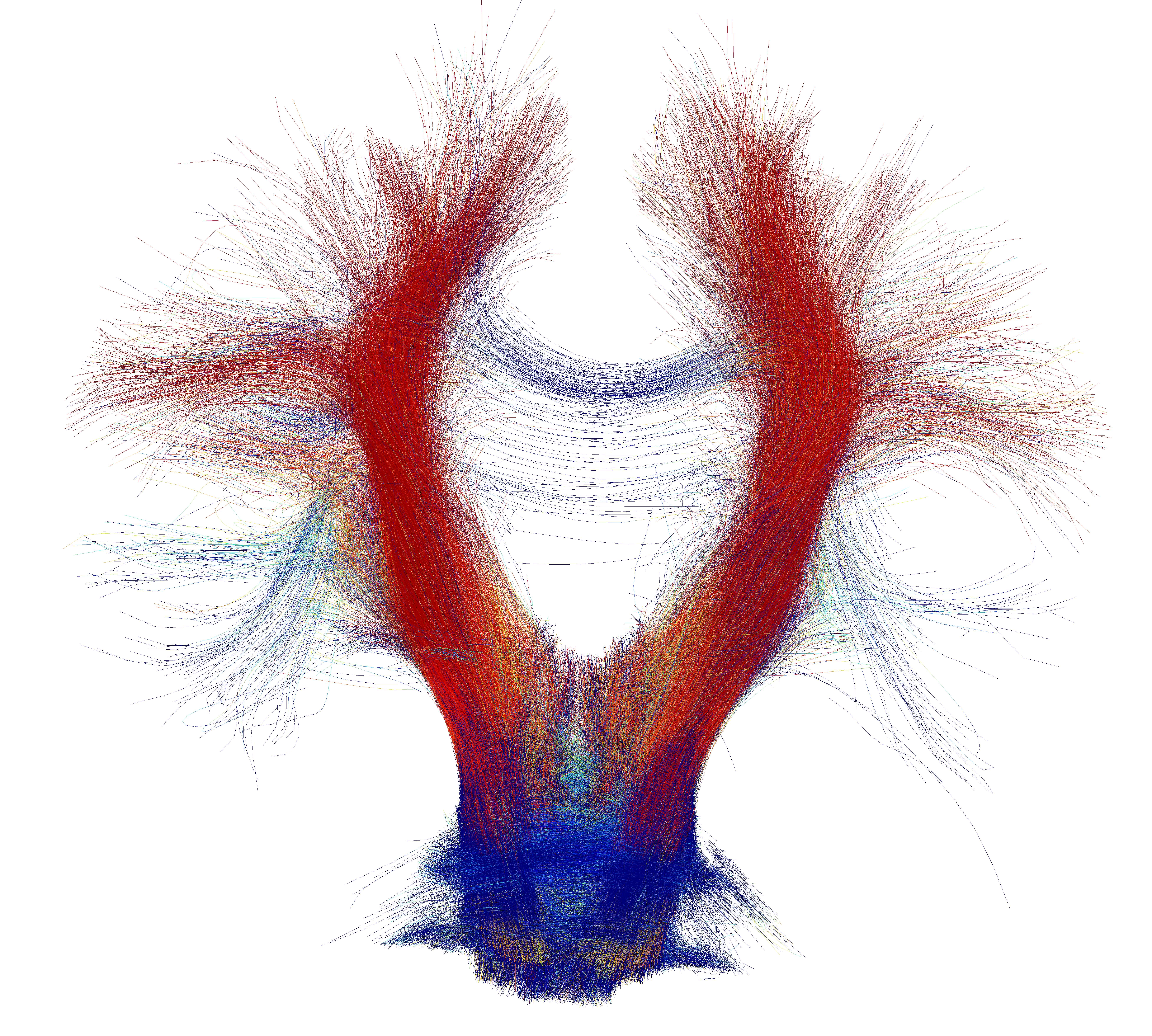} }}%
    \caption{(a)  {\em TractOracle}: an RL system in which the environment sends streamlines to TractOracle-Net as they are being tracked to score their anatomical plausibility. The scores are used to reward the agent and stop the tracking process when streamlines diverge into an implausible shape. TractOracle-RL then uses the reward function to predict anatomically-informed tractograms. (b) TractOracle-Net scores along valid and invalid cortico-spinal tracts. Streamlines correctly terminating in the motor cortex get a high plausibility score ({\color{red} red}); implausible streamlines diverging towards the corpus callosum get a low ({\color{blue} blue}) score.} %\vspace{-0.5cm}}%
    \label{fig:tractoracle_tracktolearn}%
\end{figure}

Tractography is the process of virtually reconstructing the major white matter~(WM) pathways of the human brain using diffusion magnetic resonance imaging~(dMRI)~\cite{basser2000vivo}.  Tractography has been successfully used in a number of applications such as presurgical planning~\cite{essayed2017white}, connectomics~\cite{sotiropoulos2019building}, tractometry~\cite{colby2012along} and disease progression modelling~\cite{raj2012network}.

Despite many advances~\cite{descoteaux2007regularized,tournier2019mrtrix3,girard2014towards,st2018surface}, tractography still suffers from unresolved issues in the presence of complex fiber configurations~\cite{maier2017challenge,rheault2020common}. Tractography has been described as an \emph{ill-posed problem} as it tries to infer global structure only from local information~\cite{maier2017challenge}. One common solution to this problem is to reconstruct overcomplete tractograms by seeding aggressively and then use a filtering tool, such as SIFT~\cite{smith2013sift} or COMMIT~\cite{daducci2014commit} to remove unnecessary streamlines. However, this approach suffers from several limitations as it : (1) cannot recover \emph{false negative} streamlines (i.e missing connections), (2) necessitates more data handling and extra processing power and often (3) do not eliminate \emph{false positive} streamlines completely as filtering algorithms are not immune to error~\cite{jorgens2021challenges}.

Supervised machine learning approaches have been proposed to tackle this problem~\cite{benou2019deeptract,neher2017fiber,poulin2017learn,wegmayr2021entrack} by training on curated sets of streamlines. However, datasets suitable for learning the tractography procedure are still few~\cite{poulin2022tractoinferno} and mostly limited to {\em in-silico} phantoms~\cite{cote2013tractometer,maier2017challenge}.

Reinforcement learning~(RL)-based tractography~\cite{theberge2021track,theberge2024matters} has been proposed as a way to learn tractography algorithms without reference streamlines. Instead, a reward function is used to guide the learning procedure to follow the orientation of local dMRI signal, as would typical tractography do. RL-based tractography agents tend to exhibit higher accuracy than classical tractography~\cite{theberge2024matters}.
However, the proposed reward functions do not encapsulate anatomical priors and are thus prone to "reward hacking" issues~\cite{theberge2021track,Skalse2022}.

Here we propose a new RL tractography method trained to track and filter streamlines simultaneously.  At the core of our system is a transformer network that predicts streamlines' anatomical score.  This neural net acts as an oracle in our RL framework to reward streamlines based on the anatomical plausibility of their shape, guiding agent towards reconstructing streamlines with greater accuracy during training.  Moreover, computing the plausibility score of a streamline while it is being tracked allows the system to stop the tracking when it diverges to an implausible path, preventing their reconstruction.  This system is unique and exhibits state-of-the-art results in both tractogram segmentation and tractogram generation, outperforming competitive methods.

Code~\footnote{\url{https://github.com/scil-vital/TractOracleNet}}\footnote{\url{https://github.com/scil-vital/TrackToLearn/}} is available online.

\section{Preliminaries and proposed method}

\subsubsection*{Tractography} dMRI allows for virtual reconstruction of the WM pathways through a process called tractography. From an initial 3D point $p_0$ called a \emph{seed}, a direction $\vec{u}_0$ of amplitude $\Delta$ is chosen according to a local model $v(p_0)$ (e.g. a fiber orientation distribution function, fODF~\cite{descoteaux2007regularized}) of the diffusion signal at $p_0$. Taking a step in $\Delta u_0$ gives rise to $p_1$, a new position in the WM. The process is repeated until a stopping criterion is met, i.e. exiting the WM mask. Formally:
\begin{equation}
    p_{t+1} = p_{t} + \Delta \vec{u_t},\;\;\;\; u_t \sim v(p_t) .
\end{equation}
The resulting ordered set of points $P = \{p_0, ... p_T \}$ with $T$ being the length of the sequence is called a streamline, or tract. An ensemble of streamlines is called a tractogram. Seed points can be generated throughout the WM mask to ensure proper coverage of the WM, but the resulting streamlines may overrepresent major WM pathways~\cite{jeurissen2019diffusion}. Alternatively, seed points can be generated at the WM/grey matter~(GM) interface, where axons are known to originate.

\subsubsection*{Reinforcement learning (RL)}

Tractography can be formulated as an RL learning problem~\cite{theberge2021track}.  In that case, the tractography system involves an agent $\pi$ trained to predict tracking steps by interacting with its environment, i.e. 3D dMRI signal.  RL uses the principles of the Markov Decision Process, describing the environment in terms of $S$ the set of all possible states, $A$ the set of all possible actions, $p(s_{t+1}|s_t,a_t)$ the transition probability between two consecutive states and a reward function $r(s_t,a_t)$ ($r_t$ in short).  An action $a_t \in A$ taken by the agent at state $s_t \in S$ leads to a new state $s_{t+1} \in S$ and a reward $r_t$ given by the environment.  By repeating the process of taking actions in states, a series of states and actions called a {\em trajectory} is generated.  In tractography, a state $s_t$ correspond to the dMRI signal in the vicinity of $p_t$, an action $a_t$ corresponds to the translation vector $\vec u_t$ between $p_t$ and $p_{t+1}$, and a trajectory amounts to a WM streamline.

The goal of the agent is to optimize its policy $\pi$ so as to maximize the expected discounted sum of future rewards, or return in short:
\begin{equation}
G_t = \sum_{k=t..T}[\gamma^k r(s_{t+k},a_{t+k})] \;
\end{equation}
where $\gamma$ is a discount factor that prioritizes immediate rewards over distant ones.
Central to RL policies are the concepts of value function $V_\pi(s_t)$ and Q function $Q_\pi(s_t,a_t)$. The value function estimates the expected return from a state $s_t$ and subsequent trajectories under a policy $\pi$: 
\begin{equation}
V_\pi(s) = \mathbb{E}_{s \sim \pi} \left[G_t | s_t = s\right].
\end{equation}

The Q function $Q_\pi(s_t,a_t)$ evaluates the expected return from taking an action $a_t$ in state $s_t$ and then following policy $\pi$, providing a direct measure for assessing the immediate and subsequent value of actions:
\begin{equation}
Q_\pi(s, a) = \mathbb{E}_\pi \left[r(s_t, a_t) + G_{t+1} | s_t = s, a_t = a\right].
\end{equation}

Previous RL tractography methods~\cite{theberge2024matters} are limited by their reward function $r(s_t,a_t)$ which considers only local information from the dMRI signal surrounding $s_t$.  We instead reformulate the reward function to include a factor based on the anatomical plausibility of the reconstructed streamline from $s_0$ to $s_t$.  This score is given by a neural network called {\em TractOracle-Net}.

\subsubsection*{TractOracle-Net} 
TractOracle-Net is a transformer network~\cite{devlin2018bert} which takes streamlines as input and outputs scores related to their anatomical plausibility. Figure~\ref{fig:tractoracle_tracktolearn} illustrates how TractOracle-Net interacts with the environment to score streamlines.  In the upcoming sections, it is represented as $\Omega_\psi$ where $\psi$ are the network parameters.

Its input comprises the direction between the coordinates of a streamline that is resampled to a fixed number of points, to which a "SCORE" token is prepended. The transformer network consists of a 32-dimension embedding, positional encoding, 4 transformer encoder blocs each with 4 attention heads, and a linear and sigmoid layer to convert the SCORE token to a scalar between 0 and 1, for a total of 550k trainable parameters. TractOracle-Net is trained to perform regression using the mean-squared error as a loss function. For data augmentation, streamlines are randomly flipped, randomly cut (and then re-resampled to 128 points) and Gaussian noise is applied point-wise to the streamlines. We use 0.5 as a plausible/implausible classification threshold.

\subsubsection*{TractOracle-RL}\label{sec:proposed_tracktolearn}

For TractOracle, the state $s_t$ contains the fODFs at position $p_t$ (i.e. the local diffusion orientation $v(p_t)$) to which we append 
the six surrounding fODFs and the previous 100 tracking directions~\cite{theberge2024matters}. The initial states $s_0$ are selected at the WM/GM interface, and the actions $a_t$ produced by the agents are posed as the tracking directions $\vec{u_t}$, which the environment rescales to $\Delta$ to propagate streamlines.

The reward function is made of two terms:  a local reward which accounts for how much the streamline segment at state $s_t$ is aligned with the surrounding fODFs and its previous segment, and an anatomical score of the reconstructed streamline predicted by TractOracle-Net.  Formally: 
\begin{eqnarray}
    r_t \, = \, 
    \underbrace{
    \big(|\max_{\overline{v(p_t)}}{\langle \overline{v(p_t)}, {a}_{t} \rangle| \cdot \langle a_t, a_{t-1}\rangle}\big)}_{local} + 
    \alpha 
    \underbrace{\mathds{1}_{\Omega_\psi}(P_{0..t})}_{anatomical},
\end{eqnarray}
with $\overline{v(p_t)}$ the maxima of $v(p_t)$ and $\mathds{1}_{\Omega_\psi}$ is an indicator function of the score given by TractOracle-Net :
\begin{equation}
    \mathds{1}_{\Omega_\psi}(P_{0..t}) = \begin{cases}
        1 & \text{if \;} \Omega_\psi(P_{0..t}) >= 0.5 \text{\; and \;} t = T \\
        0 & \text{else.}
    \end{cases}
\end{equation}

Moreover, we propose a new stopping criterion based on the prediction of $\Omega_\psi$:  if the score given to a streamline after a number of steps $t$ falls below $0.5$, the tracking stops. The stopping criteria become: 
\begin{equation}
     t = T \begin{cases}
        \text{if \;} \Omega_\psi(P_{0..t}) < 0.5 \text{\; and \;} t > T_{min} \\
        \text{if \;} \mathds{1}_{WM}(p_t) < 0.1 \\
        \text{if \;} (180/\pi)\langle \vec{u}_t, \vec{u}_{t-1}\rangle < \epsilon, \\
    \end{cases}   
\end{equation}
where $T_{min}$ indicates the minimum number of steps before the criterion can be enforced, $\mathds{1}_{WM}(p_t)$ is the value found via trilinear interpolation of the WM mask at position $p_t$ and $\epsilon$ indicates the maximum angle (in degrees) between two streamline segments.

\paragraph{Implementation} We set $\alpha = 10$, $T_{min} = 20$ and $\epsilon = 30$.  We use the Soft-Actor Critic~\cite{haarnoja2018soft} RL algorithm to train the agents.  The actor and twin critics are 3 layers fully-connected neural networks with a width of 1024.  In our case, $\pi$ outputs the mean and the standard deviation of a 3D Gaussian distribution which is sampled to obtain $a_t$, the tracking step. We set the learning rate and discount factor to 0.0005 and 0.95, respectively. 

\section{Experiments and results}

\subsection{Datasets and tools}\label{sec:dataset_inferno}

%We present the datasets used to train and assess the performances of TractOracle-RL and TractOracle-Net as well as the streamlines used to train the latter.

\subsubsection*{ISMRM2015}

A synthetic dataset derived from global tractography on one subject of the Human Connectome Project~\cite{maier2017challenge,fritzsche2012mitk,glasser2016human}. We performed tracking using an ensemble of five algorithms~\cite{girard2014towards,descoteaux2007regularized,tournier2019mrtrix3}, each generating 100k streamlines from both WM and WM/GM interface seeding. All other parameters were left to their respective default values. The resulting streamlines were segmented by the Tractometer~\cite{renauld2023validate} to obtain positive and negative examples.% The resulting streamlines were split into training, validation and testing sets following a 70\%/20\%/10\% regime.

\subsubsection*{BIL\&GIN}

The Brain Imaging of Lateralization by the Groupe d’Imagerie Neurofonctionnelle~(BIL\&GIN) dataset is a publicly available dataset of 453 healthy adults participants. To train TractOracle-Net, we follow the experimental procedure as described in~\cite{legarreta2021filtering} to obtain the callosal fibres of 39 randomly selected subjects from the dataset. %Following~\cite{legarreta2021filtering}, we split the subjects into training, validation and test sets following a 60\%/20\%/20\% regime.

\subsubsection*{TractoInferno}

Multi-site and multi-protocol dataset totalling 284 subjects. Because the reference bundles from~\cite{poulin2022tractoinferno} do not provide negative examples, we repeat the tracking procedure as presented in~\cite{poulin2022tractoinferno} to generate data for TractOracle-Net. We use Recobundles~\cite{garyfallidis2018recognition} to extract reference bundles and keep the unrecognized streamlines as negative examples, from which we select at random, per subject, 500k streamlines from recognized and unrecognized fibers respectively.

% \subsubsection*{Penthera 3T}
% 
% The Penthera 3T~\cite{paquette2019penthera} dataset is a collection of 13 subjects scanned six times over three sessions. All subjects were processed then using TractoFlow. We do not require streamlines for this dataset.

\subsection{TractOracle-Net performance}\label{sec:tractoracle-net-results}

\begin{table}[!tp] \centering
\caption{Classification metrics on the ISMRM2015 dataset. FINTA results are reported from~\cite{legarreta2021filtering}. Best results per metric are highlighted in \bf{bold}.}
\label{tab:tractoracle_ismrm2015}
\begin{tabular}{l   c   c   c   c }
& {Accuracy (\%)} & {Sensitivity}  & {Precision} & {F1-score} \\ \hline
{Recobundles}  & 0.91 & 0.81 & \bf{0.97} & 0.88 \\
{FINTA}         & 0.91 & 0.91 & 0.91 & 0.91 \\
{TractOracle-Net}      & \bf{0.97} & \bf{0.98} & 0.94 & \bf{0.96} \\
\end{tabular}   

\caption{Classification metrics (Mean $\pm$ stdev) on 8 test subjects of BIL\&GIN. Recobundles and FINTA results are reported from~\cite{legarreta2021filtering}. Best results are highlighted in \bf{bold}.}
\label{tab:tractoracle_bilgin}
\begin{tabular}{l   c   c   c   c }
& {Accuracy } & {Sensitivity}  & {Precision} & {F1-score} \\ \hline
 {Recobundles}         & 0.82 $\pm$ 0.03       & 0.80 $\pm$ 0.03      & 0.67 $\pm$ 0.01      & 0.70 $\pm$ 0.02 \\
{FINTA}                & 0.91 $\pm$ 0.01       & 0.91 $\pm$ 0.01      & \bf{0.78 $\pm$ 0.01} & \bf{0.83 $\pm$ 0.01} \\
{TractOracle-Net}      & \bf{ 0.96 $\pm$ 0.01} & \bf{0.92 $\pm$ 0.02} & 0.76 $\pm$ 0.02      & \bf{0.83 $\pm$ 0.02} \\
 
\end{tabular}       

\end{table}  

To ensure that TractOracle-Net can properly guide the learning procedure of RL agents and validate the chosen architecture, we first need to ensure that it can classify streamlines based on their shape. We train two instances of TractOracle-Net on the ISMRM2015 and BIL\&GIN datasets~(c.f. section~\ref{sec:dataset_inferno}) for 50 and 200 epochs, respectively. We compare ourselves to the FINTA~\cite{jorgens2021challenges} streamline filtering algorithm and Recobundles~\cite{garyfallidis2018recognition} clustering algorithm. Recobundles uses the ground-truth ISMRM2015 bundles as atlas on the dataset.

Tables~\ref{tab:tractoracle_ismrm2015} and~\ref{tab:tractoracle_bilgin} report results for this experiment. Recobundles and FINTA obtain the highest precision on ISMRM2015 and BIL\&GIN, respectively, but RecoBundles has access to ground truth bundles as opposed to the other methods. Besides precision, TractOracle outperforms Recobundles and FINTA according to all other metrics on the ISMRM2015 and BIL\&GIN datasets. 

\subsection{TractOracle {\em in-silico} performance}\label{sec:tractoracle-rl-ismrm2015}

\begin{table}[!tp] 
\centering
\caption{Tractometer scores (mean $\pm$ stddev) on ISMRM2015 for TractOracle-RL (Proposed), Track-to-Learn, and classical tractography algorithms. Scores in \textbf{bold} indicate the best method for each metric, scores in {\textcolor{black}{\textbf{red}}} indicate the method is severely outperforming the others.}
\label{tab:agents_ismrm}
\begin{tabular}{l   c   c   c   c  }
& {VC $\% \uparrow$} & {VB (/21) $\uparrow$}  & {IC $\% \downarrow$} & {IB $\downarrow$} \\ \hline

{sd\_stream} & 55.96 $\pm$ 0.21 & 19.00 $\pm$ 0.00 & 44.04 $\pm$ 0.21 & 199.80 $\pm$ 4.26  \\
{ifod2} &  31.53 $\pm$ 0.20 & 19.00 $\pm$ 0.00 & 68.47 $\pm$ 0.20 & 281.00 $\pm$ 4.00  \\
{Track-to-Learn} & 66.13 $\pm$ 1.15 & \bf{20.00 $\pm$ 0.63} & 33.87 $\pm$ 1.15 & 293.40 $\pm$ 11.8  \\
{TractOracle} & \bf{\color{black}88.05 $\pm$ 0.35} & 19.33 $\pm$ 0.47 & \bf{\color{black}11.95 $\pm$ 0.35} & \bf{195.67 $\pm$ 4.99}  \\
\hline
& {OL $\% \uparrow$} & {OR $\% \downarrow$} & {F1 $\% \uparrow$} & {NC $\downarrow$} \\ \hline   
{sd\_stream} & 38.85 $\pm$ 0.05 & 3.59 $\pm$ 0.05 & 52.47 $\pm$ 0.03 & 9.36 $\pm$ 0.18 \\
{ifod2} & 48.70 $\pm$ 0.11 & \bf{8.18 $\pm$ 0.19} & \bf{59.10 $\pm$ 0.07} & 12.45 $\pm$ 0.08  \\

{Track-to-Learn} & \bf{53.84 $\pm$ 2.28} & 29.94 $\pm$ 2.00 & 57.43 $\pm$ 1.84 & 2.85 $\pm$ 0.42  \\
{TractOracle} & 48.43 $\pm$ 0.64 & 17.68 $\pm$ 1.00 & 57.00 $\pm$ 0.47 & \bf{\color{black}0.73 $\pm$ 0.12}  \\
\hline
\end{tabular}       
\end{table}  
We then investigate the improvements brought by TractOracle-RL over other tractography methods by training and tracking on the ISMRM2015 dataset. All agents were trained for 1000 epochs, five times with different random seeds. We used the TractOracle-Net model trained on the ISMRM2015 dataset from the previous experiment. We compare the proposed method against some of the most widely used algorithms: the sd\_stream and ifod2 algorithms from MRTrix3~\cite{tournier2019mrtrix3} as well as Track-to-Learn~\cite{theberge2024matters}. For all methods, we track five times at 20 seeds per voxels at the WM/GM interface using five different random seeds and report mean Tractometer metrics.

Table~\ref{tab:agents_ismrm} reports Tractometer metrics for all agents considered in this experiment.  As shown in red, TractOracle obtains by far the highest Valid Connection (VC) rate ($22\%$ better than the 2nd best method), as well as the lowest Invalid Connection~(IC) (3x lower than the 2nd best) and No-connection~(NC) rates (3x lower than the 2nd best). While the proposed method reduces the overlap~(OL) of recovered bundles compared to Track-to-Learn, it also reduces the overreach~(OR) preserving a similar F1 score, as well as greatly reducing the number of invalid bundles~(IB) reconstructed. Compared to classical tractography, the proposed method has a similar number of VB and OL and F1 rates while having significantly higher VC rates and significantly lower IB numbers, IC, NC rates, the best scores ever reported on this dataset.

\subsection{TractOracle {\em in-vivo} performance}\label{sec:tractoracle-rl-invivo}

\begin{figure}[tb]%
    \centering
    \includegraphics[width=\linewidth]{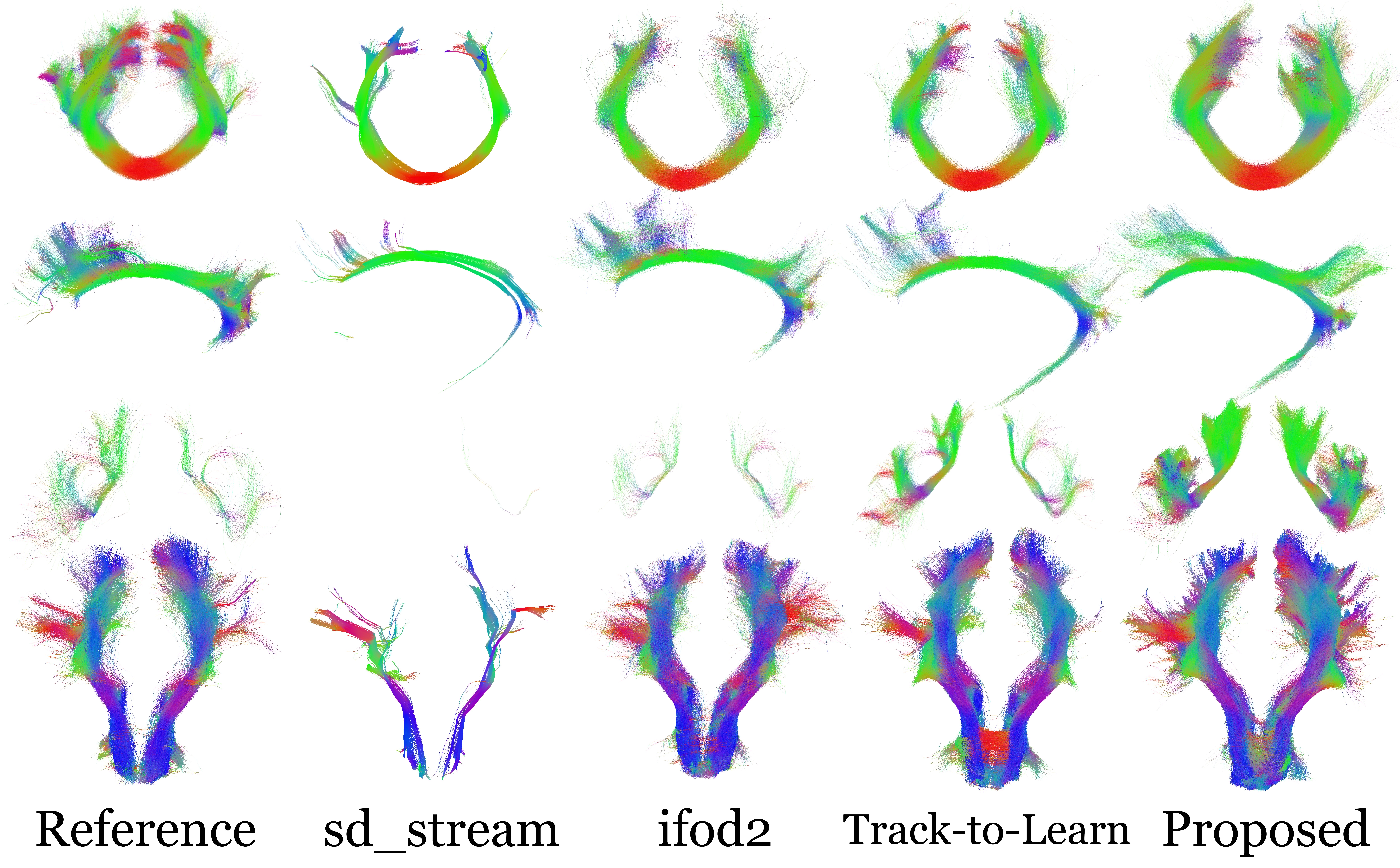} 
    \caption{Visualization of the occipital part of the corpus callosum (1st row), the left cingulum (2nd row), the left-right uncinate fasciculus (3rd row) and the left-right parieto-occipito pontine tracts (4th row) from subject 1006 of the Tractoinferno dataset (reference) and as reconstructed by all methods considered in this work.}%
    \label{fig:tractoinferno_bundles}%
\end{figure}

After validating the {\em in-silico} performance of our agents, we turn  to {\em in-vivo} subjects. We train TractOracle-Net on the reference streamlines of TractoInferno as described in section~\ref{sec:dataset_inferno} for 10 epochs. We report an accuracy of 89.21\% on the TractoInferno test split. We then train TractOracle-RL agents on the TractoInferno dataset and track on the test set, repeating the experimental procedure and reusing the same hyperparameters from the previous experiment. 

We segment the tractograms using Recobundles to provide a qualitative comparison to reference bundles of the test set. We additionally evaluate the anatomical accuracy of the reconstructed tractograms by classifying them using Recobundles~\cite{garyfallidis2018recognition}, TractOracle-Net and extractor\_flow~\cite{petit2023structural}, an automated set of anatomical rules derived from current anatomical knowledge, and report the number of streamlines that were classified as plausible by the tools.

Figure~\ref{fig:tractoinferno_bundles} displays some bundles from subject 1006 of the dataset and as reconstructed by all methods. We can observe that the proposed method reconstructs highly visually appealing bundles, with excellent fanning. Moreover, the proposed method tends to reconstruct parts of the bundles that even the reference dataset misses, such as the "tail" of the cingulum, most of the uncinate fasciculus and some fanning of the parieto-occipito pontine tracts. sd\_stream provides generally thin bundles, whereas ifod2 and Track-to-Learn produce similarly voluminous bundles.

\begin{table}[tp] \centering

    \caption{Total number of streamlines recovered by Recobundles~\cite{garyfallidis2018recognition}, extractor\_flow~\cite{petit2023structural} and TractOracle-Net for all tracking algorithms considered.}\label{tab:agents_tractoinferno_vs}
    \begin{tabular}{l   r  r r r }
         & {\ Recobundles $\uparrow$} & {\ extractor\_flow $\uparrow$} & {\ TractOracle-Net $\uparrow$} \\ \hline
        {sd\_stream}     & 2,713,507       & 7,698,181 & 28,901,678  \\
        {ifod2}          & 2,362,290       & 15,833,504 & 24,228,612 \\
        {Track-to-Learn} & 9,309,681       & 46,135,363 & 85,447,863 \\
        {TractOracle-RL}    & \bf{25,028,287} & \bf{55,761,074} & \bf{184,355,412} \\
        \hline
    \end{tabular}       
\end{table}  

Table~\ref{tab:agents_tractoinferno_vs} underlines that TractOracle produces highly anatomically accurate tractograms.  Despite using the same number of seeding points for all methods, the number of plausible streamlines reported by TractOracle is drastically higher than other methods regardless of the classification tool used, going from an average improvement of 7x for sd\_stream and ifod2, to 2x for Track-to-Learn.

\section{Discussion and conclusion}\label{sec:discussion_and_conclusion}

In this work, we presented TractOracle, a highly accurate tractography system which evaluates and generates streamlines at the same time. From the results above, TractOracle-Net outperforms state-of-the-art streamline classification algorithms~(c.f. tables~\ref{tab:tractoracle_ismrm2015},~\ref{tab:tractoracle_bilgin}) while TractOracle-RL produces highly accurate~(c.f. tables~\ref{tab:agents_ismrm},~\ref{tab:agents_tractoinferno_vs}) and voluminous~(c.f. figure~\ref{fig:tractoinferno_bundles}) tractograms.  As of today, TractOracle is the most effective machine learning WM tracking method. 

Future work should aim at improving the overlap between the reconstructed tractograms and reference bundles~(c.f. table~\ref{tab:agents_ismrm}). But most importantly, future work should aim at giving the tracking agent global context of the diffusion volume to finally tackle the ill-posedness of tractography.

\newpage
\bibliographystyle{unsrtnat}
\bibliography{references}

\end{document}